\documentclass[10pt, conference]{IEEEtran}

\usepackage{url}
\usepackage{array}
\usepackage{graphicx}
\usepackage{xcolor}
\usepackage{caption}
\usepackage{subcaption}
\usepackage{algorithm}
\usepackage{algorithmic}
\usepackage{amsmath}
\usepackage{cleveref}
\usepackage{tikz}
\usetikzlibrary{shapes.geometric, arrows}

\usepackage{pgfplots}

\begin{document}

\title{Time Series Classification of Supraglacial Lakes Evolution over Greenland Ice Sheet}


\author{
    \IEEEauthorblockN{
        Emam Hossain\IEEEauthorrefmark{1},
        Md Osman Gani\IEEEauthorrefmark{1},
        Devon Dunmire\IEEEauthorrefmark{2}, 
        Aneesh Subramanian\IEEEauthorrefmark{3},
        Hammad Younas\IEEEauthorrefmark{4}
    }
    \IEEEauthorblockA{
        \IEEEauthorrefmark{1}Department of Information Systems, University of Maryland Baltimore County, USA \\
        \IEEEauthorrefmark{2}Department of Earth and Environmental Sciences, KU Leuven, Belgium\\
        \IEEEauthorrefmark{3}Department of Atmospheric and Oceanic Sciences, University of Colorado Boulder, USA \\
        \IEEEauthorrefmark{4}St. John's School, Houston, USA \\
        Email: \textit{\{emamh1, mogani\}@umbc.edu, devon.dunmire@kuleuven.be, aneeshcs@colorado.edu, hammadyounas27@icloud.com}
    }
}

\maketitle
\thispagestyle{empty}

\begin{abstract}
    The Greenland Ice Sheet (GrIS) has emerged as a significant contributor to global sea level rise, primarily due to increased meltwater runoff. Supraglacial lakes, which form on the ice sheet surface during the summer months, can impact ice sheet dynamics and mass loss; thus, better understanding these lakes' seasonal evolution and dynamics is an important task. This study presents a computationally efficient time series classification approach that uses Gaussian Mixture Models (GMMs) of the Reconstructed Phase Spaces (RPSs) to identify supraglacial lakes based on their seasonal evolution: 1) those that refreeze at the end of the melt season, 2) those that drain during the melt season, and 3) those that become buried, remaining liquid insulated a few meters beneath the surface. Our approach uses time series data from the Sentinel-1 and Sentinel-2 satellites, which utilize microwave and visible radiation, respectively. Evaluated on a GrIS-wide dataset, the RPS-GMM model, trained on a single representative sample per class, achieves 85.46\% accuracy with Sentinel-1 data alone and 89.70\% with combined Sentinel-1 and Sentinel-2 data. This performance significantly surpasses existing machine learning and deep learning models which require a large training data. The results demonstrate the robustness of the RPS-GMM model in capturing the complex temporal dynamics of supraglacial lakes with minimal training data.
\end{abstract}

\section{Introduction}\label{sec_intro}

\begin{figure}[!h]
    \centering
    \includegraphics[width=0.48\textwidth]{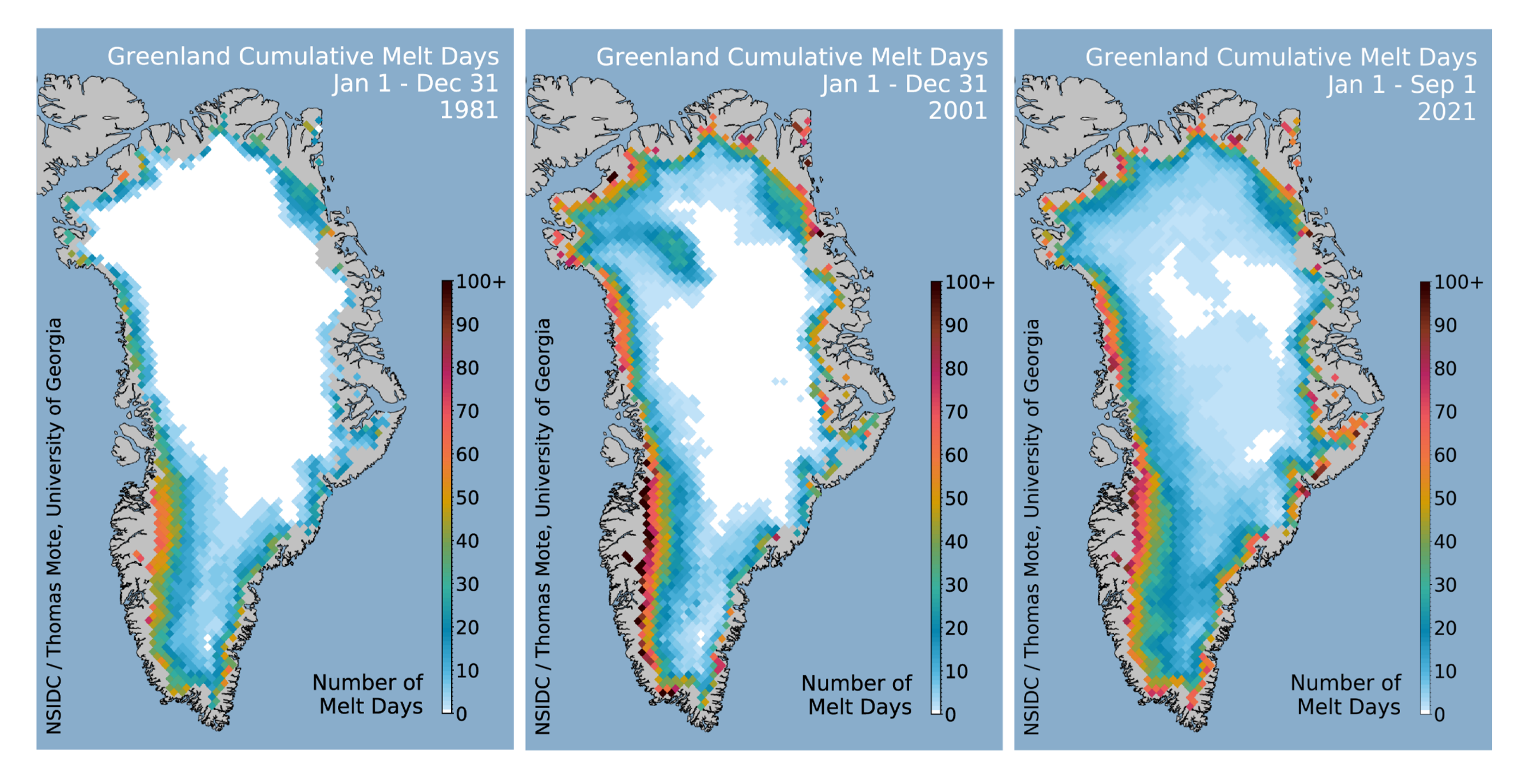}
    \caption{Cumulative melting days over the GrIS in 1981, 2001, and 2021. Source: National Snow and Ice Data Center (NSIDC) \cite{nsidc2021greenland}}
    \label{fig:gris_melting}
\end{figure}

The Greenland Ice Sheet (GrIS) contributes substantially to rising sea levels and can raise the sea level by more than 7 meters if melted entirely \cite{smith2020}. The GrIS has been losing mass annually, and since 1992, it has been estimated to contribute nearly 14~mm to global sea level rise \cite{otosaka2023mass}. The ice sheet loses mass through dynamic (speed-up of ice flow) and surface (meltwater runoff) processes. Recent studies have indicated that surface and meltwater processes have become the predominant contributor to GrIS mass loss \cite{vanDen'16}, highlighting the increasingly important role of GrIS surface melt. \Cref{fig:gris_melting} illustrates the growing number of cumulative melting days over the GrIS in the years 1981, 2001, and 2021, indicating a substantial increase in melting in recent decades.

Supraglacial lakes are meltwater features that form on the ice sheet surface during the summer months. Understanding how these features evolve throughout the melt season is important because of their potential impact on ice sheet dynamics and mass loss \cite{chu2014greenland}. Several things can happen to supraglacial lakes throughout a melt season. Many lakes refreeze at the end of the melt season as temperatures drop below 0$^{\circ}C$. This refreezing creates impermeable ice layers, alters firn (the partially compacted snow layer) density, reduces firn air content, and affects future meltwater percolation and storage \cite{machguth2016greenland, macferrin2019rapid}. Some lakes, however, do not refreeze entirely and remain liquid buried a few meters underneath the ice sheet surface \cite{dunmire2021contrasting, koenig2015wintertime}. These features, called buried lakes, may temporarily store meltwater and reduce immediate runoff and mass loss, but their long-term impact on the ice sheet is still relatively unknown. Numerous supraglacial lakes also drain throughout the melt season, sometimes slowly via overflow drainage and sometimes rapidly via hydrofracture. Hydrofracture occurs when meltwater activates or extends fractures in the ice \cite{das2008fracture}, and creates hydrologic pathways from the ice sheet surface to the bedrock, creating a means for surface meltwater to impact basal friction and ice velocity \cite{zwally2002surface, hoffman2011links}.


Numerous studies have used optical and near-infrared imagery from satellites such as Landsat and Sentinel-2 to identify and monitor supraglacial lakes and channels \cite{miles2017toward, williamson2018dual, hu2021distribution, domgaard2024altimetry}. Other works have taken advantage of microwave imagery, which overcomes some of the limitations of optical imagery during the polar night and cloudy conditions. Further, microwave imagery can be used to detect and monitor buried lakes \cite{dunmire2021contrasting, benedek2021winter, schroder2020perennial}, as microwaves can penetrate several meters beneath the ice surface \cite{rignot2001penetration}. In this study, we combine both optical and microwave imagery to introduce a computationally efficient time series classification method where we fit Gaussian Mixture Models (GMMs) to Reconstructed Phase Space (RPS) of different lakes classification using Maximum Likelihood Estimation (MLE). We classify supraglacial lakes according to their seasonal changes: 1) lakes that \textit{refreeze} completely in winter, 2) those that \textit{drain} either slowly or quickly during summer, and 3) those that get \textit{buried} under ice or snow during winter. These classifications are crucial for understanding how meltwater is stored and released, influencing ice flow and sea level rise (\Cref{sec_lake_types}). Analyzing data from 777 lakes across all six sub-regions of the GrIS for the years 2018 and 2019, our approach classifies lakes using only one representative sample per class. This method is computationally more efficient and provides more accurate classifications compared to existing machine learning and deep learning models for time series classifications (\Cref{sec_tsc_models}).

\section{Related Works}

\subsection{Supraglacial Lakes}

Understanding the hydrologic processes of the GrIS and their implications for its mass balance has been a subject of significant research interest in recent years. This section provides a comprehensive overview of the related works in this domain, encompassing studies on the formation and behavior of supraglacial lakes, their impact on ice dynamics, and the methods used for their detection and monitoring.

Previous studies, such as those by \cite{smith2015efficient} and \cite{leeson2015supraglacial}, investigated the evolution of supraglacial lakes and their dependence on topographic features, highlighting the influence of bed topography on their formation and persistence. \cite{tedstone2015decadal} further explored the seasonal variability of supraglacial lakes, emphasizing their significance in understanding ice sheet dynamics. The formation and drainage of supraglacial lakes have profound implications for ice dynamics and mass loss from the GrIS. \cite{vanDen'16} demonstrated a shift in the predominant cause of mass loss from ice discharge to meltwater runoff, underscoring the increasing significance of surface melt in the ice sheet's mass balance. \cite{macferrin2019rapid} investigated the role of refrozen meltwater in altering firn properties and influencing future meltwater percolation, highlighting the complex feedback mechanisms between surface melt and ice dynamics. Furthermore, studies by \cite{chu2014greenland} and \cite{benedek2021winter} examined supraglacial lakes' seasonal variability and spatial distribution, providing insights into the factors influencing their formation and evolution. \cite{koenig2016annual} and \cite{rignot2006changes} investigated the role of supraglacial lake drainage in modulating ice sheet dynamics, highlighting the potential for abrupt changes in ice flow and mass loss.

Satellite-based and remote sensing techniques have emerged as valuable tools for detecting and monitoring supraglacial lakes on the GrIS. \cite{banwell2012modeling} and \cite{mcmillan2019sentinel} utilized multispectral satellite imagery, including data from MODIS (Moderate Resolution Imaging Spectroradiometer) and Landsat satellites, to identify supraglacial lakes and channels, enhancing the spatial coverage of surface meltwater features. \cite{hochreuther2021fully} explored the use of Sentinel-2 imagery for automated detection of supraglacial lakes, highlighting the potential for improving temporal resolution and monitoring capabilities. In addition to optical imagery, SAR imagery has been increasingly employed to detect supraglacial lakes and buried lakes on the GrIS. \cite{mcmillan2019sentinel} demonstrated the utility of SAR for detecting supraglacial lakes, while \cite{schroder2020perennial} utilized SAR imagery to identify perennial supraglacial lakes. \cite{benedek2021winter} and \cite{dunmire2021contrasting} further investigated the use of SAR for detecting and monitoring buried lakes, highlighting its capability to penetrate snow and ice.

\subsection{Machine Learning and Deep Learning} \label{sec_tsc_models}

There are several established machine learning (ML) and deep learning (DL) models widely used for time series classification tasks. Long Short-Term Memory (LSTM) networks and Fully Convolutional Networks (FCN) are among the prominent models used in this domain. Residual Networks (ResNet) and Recurrent Neural Networks (RNNs) are also integral to this field, offering unique time series analysis capabilities. The following defines some popular classification techniques that are selected based on their proven effectiveness and unique methodologies for handling time series data.

\textit{LSTMFCNClassifier:} Combines LSTM and FCN to leverage both sequential dependencies and local feature extraction.

\textit{FCNClassifier:} Utilizes FCN to capture local patterns in time series data without fully connected layers, allowing it to handle input sequences of varying lengths.

\textit{ResNetClassifier:} Employs ResNet with residual connections to facilitate the training of deep networks, effectively capturing complex patterns in the data.

\textit{SimpleRNNClassifier:} Uses RNNs with recurrent connections to capture dependencies over time, making it suitable for time series classification.

\textit{KNeighborsTimeSeriesClassifier:} A non-parametric, instance-based learning algorithm that classifies data points based on the majority class among their $k$-nearest neighbors in the feature space.

\section{Background} \label{sec_background}

This section outlines the essential components of the RPS-GMM model. We provide an overview of reconstructed phase space and Gaussian mixture models and discuss the maximum likelihood classifier and the evolution of various supraglacial lakes over time.

\subsection{Reconstructed Phase Space}

The seasonal evolution of supraglacial lakes can be viewed as a dynamic system where their state changes over time due to various environmental factors. The Reconstructed Phase Space (RPS) is particularly suitable for this research as it enables the capture of underlying dynamics from time series data, such as remote satellite observations. A dynamical system describes the temporal evolution of a system to capture its dynamics. A phase space represents all possible states of the system that evolve, and the dynamics map describes how the system evolves. The RPS captures the underlying dynamics of a system from time series observations, allowing for the analysis of complex and non-linear behaviors in supraglacial lakes. 

According to Takens' embedding theorem, a time series \( x = \{x_n\}, n = 1, \ldots, N \) can be converted into state vectors using time-delay embedding:

\begin{equation}
    X_n = [x_n, x_{n-\tau}, \ldots, x_{n-(d-1)\tau}],
\end{equation}

where $\tau$ is the time delay and $d$ is the embedding dimension \cite{takens2006detecting}. This embedding reconstructs the state and dynamics of the unknown system from observed measurements. The dimension $d$ should be greater than twice the box-counting dimension of the original system \cite{povinelli2004time}. If $d$ is unknown, it can be estimated using the false nearest-neighbor techniques, and $\tau$ can be determined by finding the first minimum of the automutual information \cite{kantz2003nonlinear}.

\subsection{Gaussian Mixture Models}

Gaussian Mixture Models (GMMs) are employed to model the distribution of dynamics represented by the RPS. The seasonal evolution of supraglacial lakes involves complex, non-linear behaviors that are well-captured by GMMs, which can model multiple underlying distributions within the data. A GMM is a weighted sum of \( M \) Gaussian distributions:

\begin{equation}
p(X \mid \lambda) = \sum_{i=1}^{M} w_i \mathcal{N}(X \mid \mu_i, \Sigma_i),
\end{equation}

where \( \lambda = \{w_i, \mu_i, \Sigma_i\} \) are the weights, means, and covariance matrices of the mixture components \cite{reynolds2009gaussian}. These parameters of the GMM are estimated using the Expectation-Maximization (EM) algorithm, which iteratively maximizes the likelihood of the data \cite{moon1996expectation}. The EM algorithm includes two main steps:

\textit{1. {Expectation (E-step):}} Calculate the responsibilities for each data point \( X_i \):

\begin{equation}
    \gamma_{ij} = \frac{w_j \mathcal{N}(X_i \mid \mu_j, \Sigma_j)}{\sum_{k=1}^{M} w_k \mathcal{N}(X_i \mid \mu_k, \Sigma_k)}.
\end{equation}

\textit{2. Maximization (M-step):} Update the parameters using the responsibilities:

\begin{equation}
    w_j = \frac{1}{N} \sum_{i=1}^{N} \gamma_{ij},
\end{equation}

\begin{equation}
    \mu_j = \frac{\sum_{i=1}^{N} \gamma_{ij} X_i}{\sum_{i=1}^{N} \gamma_{ij}},
\end{equation}

\begin{equation}
    \Sigma_j = \frac{\sum_{i=1}^{N} \gamma_{ij} (X_i - \mu_j)(X_i - \mu_j)^T}{\sum_{i=1}^{N} \gamma_{ij}}.
\end{equation}

The E-step and M-step are repeated until the parameters converge, ensuring the GMM optimally fits the data.

\subsection{Maximum Likelihood Classifier}

A Bayesian maximum likelihood classifier is used to classify the test data. For each test point \( X_k \), the likelihoods are computed for each model \( a_i \):

\begin{equation}
    p(X \mid a_i) = \prod_{k=1}^{T} p(x_k \mid a_i),
\end{equation}

where \( X = \{x_1, x_2, \ldots, x_T\} \) is the sequence of observations. By leveraging the likelihoods derived from the GMM, the classifier can effectively distinguish between the different types of lakes. After computing all the likelihoods, the class with the maximum likelihood, $\hat{a}$ (i.e., predicted class), is determined using the following equation \cite{gani2019light}:

\begin{equation}
\hat{a} = \arg\max_{i} p(X \mid a_i)
\end{equation}

\subsection{Evolution of Supraglacial Lakes} \label{sec_lake_types}

\begin{figure}[!h]
    \centering
    \includegraphics[width=0.48\textwidth]{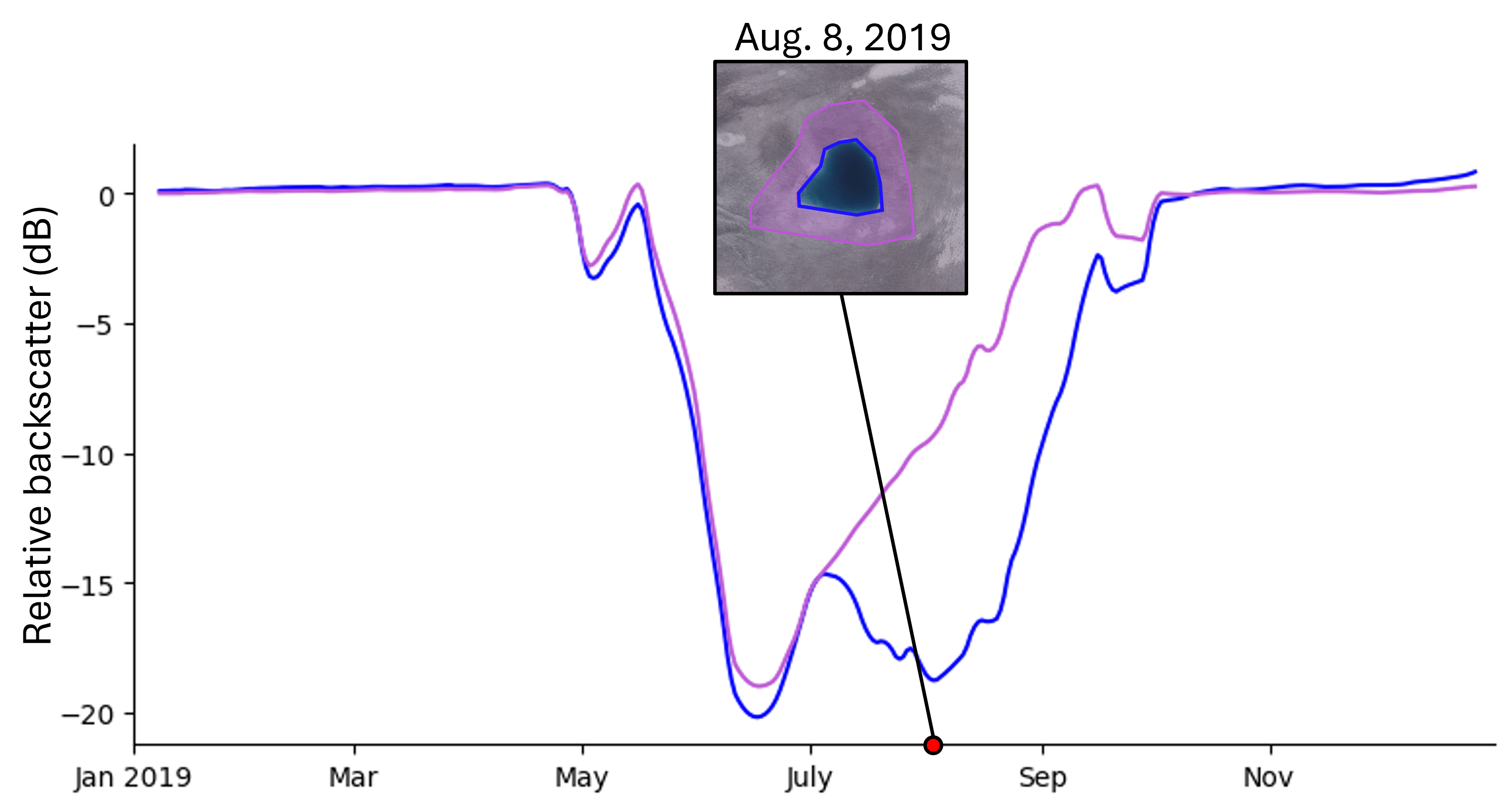}
    \caption{Comparison of backscatter signal received from within a lake (blue) and its vicinity (purple).}
    \label{fig:backscatter_life}
\end{figure}

We observe distinct evolutionary changes in supraglacial lakes from satellite data, specifically using the Sentinel-1 (S1, microwave) and Sentinel-2 (S2, optical) satellites. We classify lakes into three distinct categories: \textit{refreezing lakes}, \textit{draining lakes}, and \textit{buried lakes}. From the S1 microwave imagery, we use the horizontally-transmitted, vertically-received (HV) band, previously used for buried lake detection \cite{dunmire2021contrasting, benedek2021winter}. For each S1 image, we calculate the average HV value both within the lake outline and in the immediate vicinity outside the lake bounds, within a 750$\sim$m buffer of the lake. \Cref{fig:backscatter_life} illustrates an average backscatter timeseries within the lake bounds (blue line, $HV_{lake}$) and from the area surrounding the lake (purple line, $HV_{background}$). By differencing these two backscatter signals using \Cref{equ:hv_anom}, we can identify the backscatter anomaly for a given lake, denoted as $HV_{anom}$. 

\begin{figure*}[!h]
    \centering
    \begin{subfigure}[t]{0.33\textwidth}
        \centering
        \includegraphics[width=\textwidth]{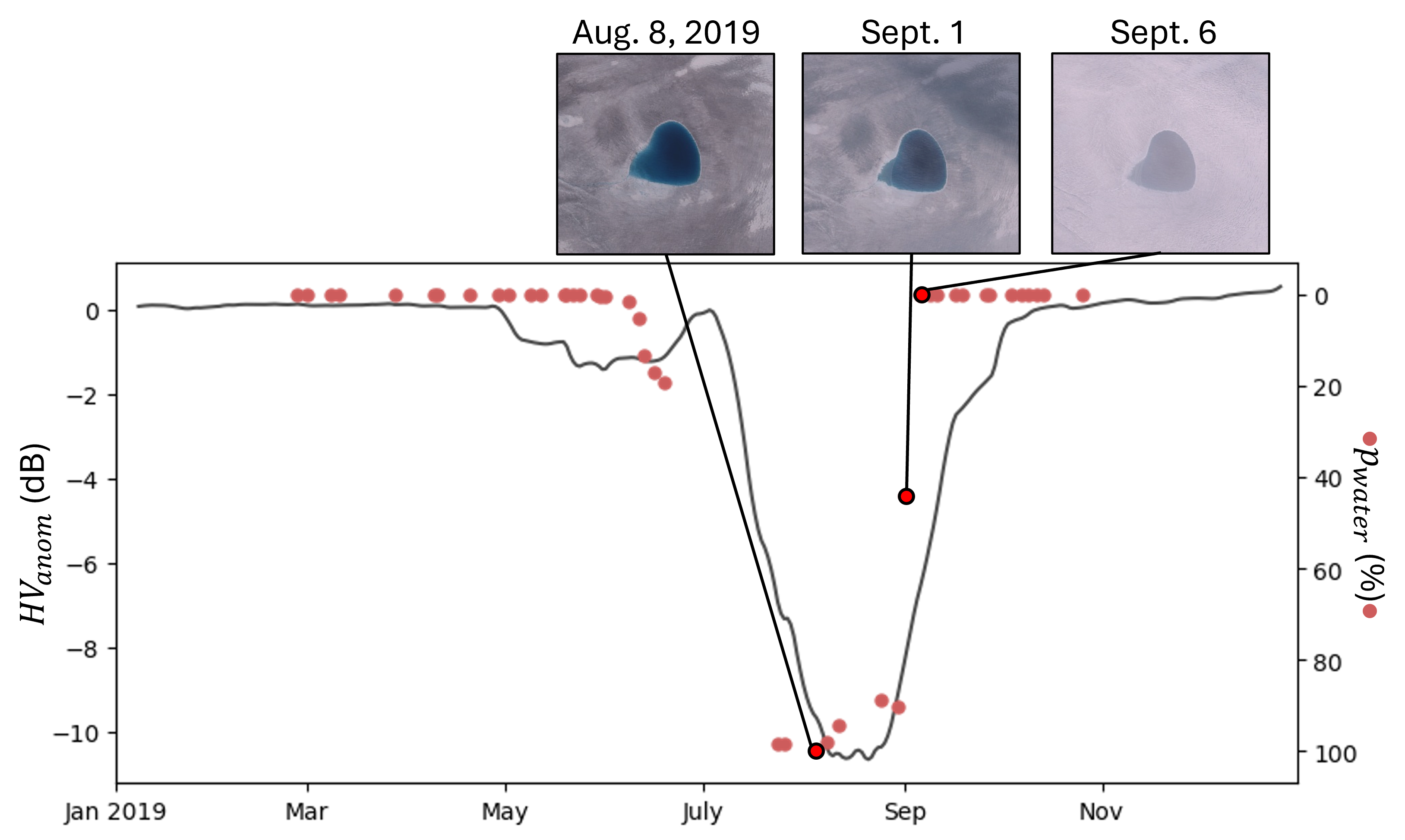}
        \caption{Refreeze lake}
        \label{fig:refreeze}
    \end{subfigure}%
    \hfill
    \begin{subfigure}[t]{0.33\textwidth}
        \centering
        \includegraphics[width=\textwidth]{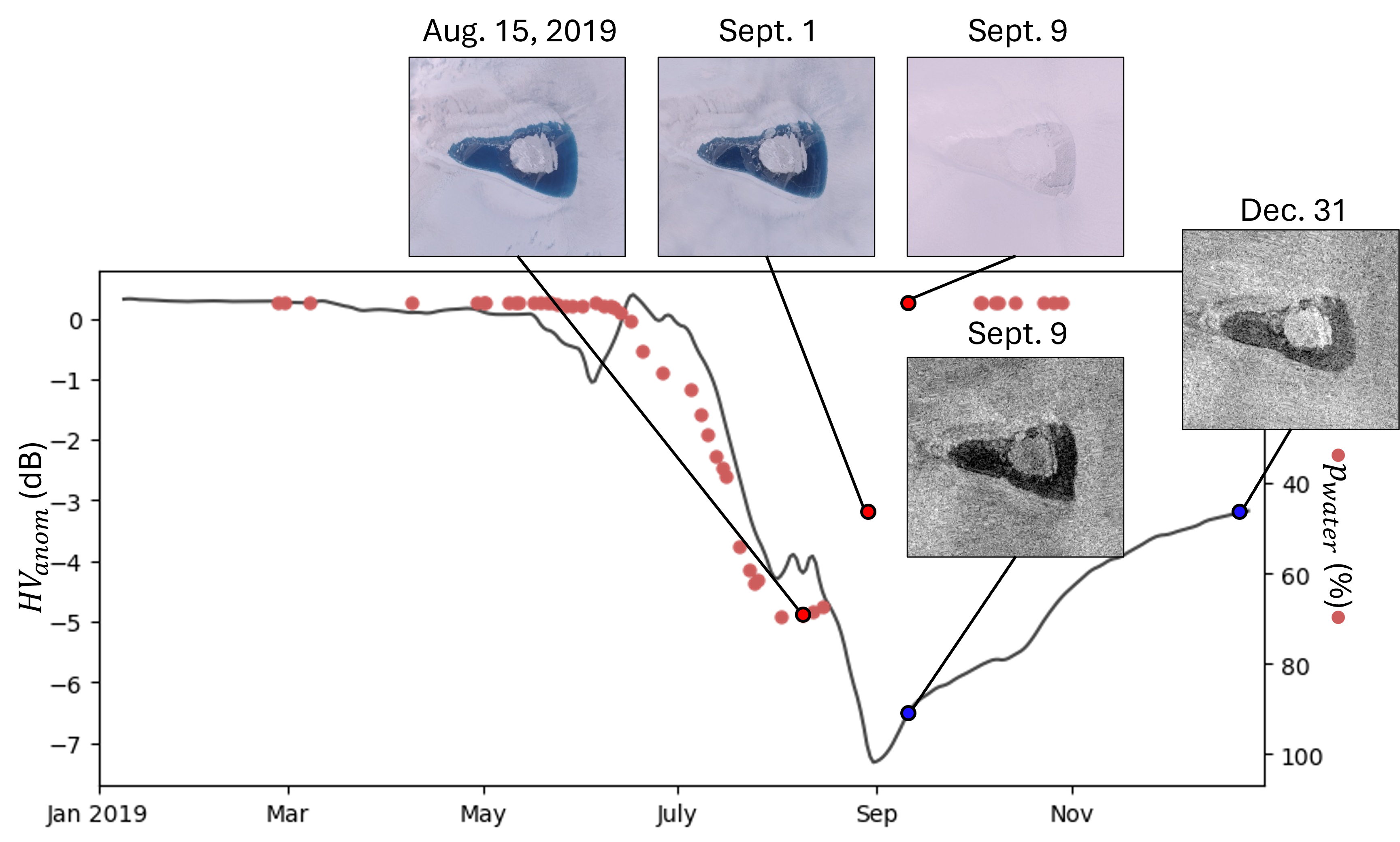}
        \caption{Buried lake}
        \label{fig:buried}
    \end{subfigure}%
    \hfill
    \begin{subfigure}[t]{0.33\textwidth}
        \centering
        \includegraphics[width=\textwidth]{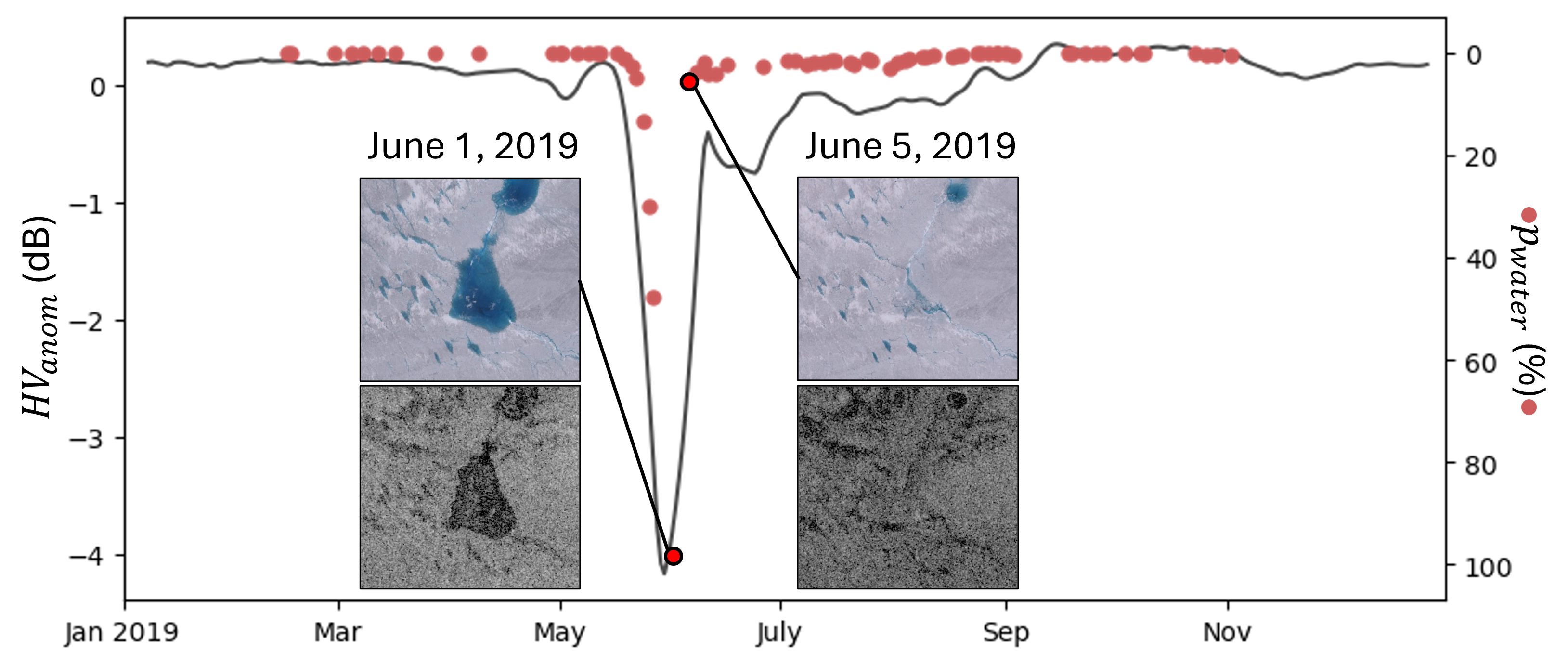}
        \caption{Drained lake}
        \label{fig:drain}
    \end{subfigure}
    \caption{Evolution of different types of supraglacial lakes over time. The grey line shows the time series of backscatter differences (left y-axis) and the red dots represent the percentage of water coverage from each S2 image observation (right y-axis). Black-and-white and color images are from Sentinel-1 and Sentinel-2 satellites, respectively.}
    \label{fig:lake_types}
\end{figure*}

We also obtain a time series with information from optical imagery for each lake. From each S2 optical image, we determine the water percentage ($p_{water}$) inside the lake by calculating the ratio of water-identified pixels to the total number of pixels within the lake. This calculation is detailed in \Cref{equ:p_water}. $N_{water}$ represents the number of pixels identified as water in the S2 image, and $N_{total}$ is the total number of pixels within the lake. \Cref{fig:lake_types} illustrates an example time series for all three lake types.


\begin{equation}
    HV_{anom} = HV_{lake} - HV_{background}
    \label{equ:hv_anom}
\end{equation}

\begin{equation}
    p_{water} = \frac{N_{water}}{N_{total}} \times 100\%
    \label{equ:p_water}
\end{equation}

\textit{Refreeze lakes} form when temperatures decrease towards the end of the melt season, causing the lakes to transition from liquid to frozen. \Cref{fig:refreeze} illustrates that both $HV_{anom}$ and $p_{water}$ decline at the end of the summer and approach zero in the fall, indicating the lake water has fully refrozen.

The optical timeseries for a \textit{buried lake}  appears similar to that of a \textit{refreeze lake}, as $p_{water}$ decreases to zero at the end of the melt season. However, microwave images and the backscatter $HV_{anom}$ timeseries in \Cref{fig:buried} indicate that some liquid water remains buried beneath the surface, even at the end of the year. Water strongly absorbs microwave radiation and thus areas with liquid water presence appear relatively dark in the microwave imagery.

\textit{Drained lakes} undergo a significant reduction in water volume, typically marked by a sharp decline in $p_{water}$. We classify lakes as draining if the lake water is drained to the ice bed and $p_{water}$ approaches zero during the melt season. \Cref{fig:drain} depicts the optical and microwave timeseries of a lake experiencing rapid drainage, with imagery indicating that the lake was drained over the four days between June 01-05.

\section{Experiments} \label{sec_methodology}

\subsection{Data Collection and Preprocessing}
For this study, we utilize a comprehensive pan-Greenland dataset \cite{dunmire2021contrasting, GrIS_BL_dataset}, which includes detailed outlines of supraglacial lakes with a resolution of 30 meters. This dataset consists of 3,846 supraglacial lakes during the 2018 melt season and 6,146 supraglacial lakes during 2019, each with a surface area $>$ 0.05 $km^2$. The years 2018 and 2019 are chosen to capture a range of climatic conditions, with 2018 representing a cooler melt season and 2019 representing a warmer melt season \cite{dunmire2024greenland, subramanian2024fate}, providing a comprehensive understanding of supraglacial lake dynamics under different temperature regimes. For this work, we manually label the time series of 777 lakes into three classes: \textit{refreeze} (189 lakes), \textit{drained} (392 lakes), and \textit{buried} (196 lakes). The dataset spans all six subregions of the Greenland Ice Sheet: Northeast (NE), Northwest (NW), North (NO), Central West (CW), Southeast (SE), and Southwest (SW) \cite{rignot2012ice}. This dataset is selected for its ice-sheet-wide coverage and high spatial resolution.

We utilize satellite imagery from two sources: microwave imagery (S1 satellite) and optical imagery (S2 satellite). Data is acquired using the Google Earth Engine (GEE) \cite{Gorelick2017RemoteEveryone}. S1 imagery is already preprocessed within GEE by applying thermal noise removal, radiometric calibration, terrain correction, and conversion to decibels via log scaling. For each supraglacial lake identified in 2018 and 2019, the dataset includes S1 imagery from January 1 to December 31 of the respective year. For optical imagery, we use S2 Level-1C orthorectified top-of-atmosphere reflectance, specifically Bands 2 (Blue, 20~m), 3 (Green, 20~m), 4 (Red, 20~m), 10 (Cirrus, 60~m), and 11 (SWIR 1, 20~m). To generate a complete annual time series of $HV_{anom}$ and $p_{water}$ for each lake, we linearly interpolated between S1 and S2 observations and applied a 12-day smoothing filter to $HV_{anom}$ to reduce variability across different S1 orbits. Our analysis focuses on time series data from May 1 to December 31 to capture the seasonal dynamics of supraglacial lakes during the melt season.

\subsection{Methodology}

In this study, we classify supraglacial lakes into three categories: \textit{refreeze}, \textit{drain}, and \textit{buried} using Reconstructed Phase Space (RPS) and Gaussian Mixture Models (GMM). The methodology is divided into two main phases: training and testing. The training phase begins by selecting one representative sample for each class. We then construct the RPS for these samples using time lag ($\tau$) and embedding dimension ($d$). Instead of using false nearest neighbor or automutual information methods, we employ a grid search technique to select $\tau$ and $d$ from a range of [2, 30]. For each combination of $\tau$ and $d$, we construct the RPS for the representative samples of each class. We then initialize GMMs for each class with 10 mixtures. For each GMM, we use $k$-means clustering to generate 10 different sets of initial parameters and select the best-performing set. Subsequently, we train the GMM models $M = \{M_1, M_2, M_3\}$ for each class—\textit{refreeze}, \textit{drain}, and \textit{buried}—using the EM algorithm. The trained GMMs serve as probabilistic models for each class of supraglacial lakes. This approach is computationally efficient, requiring only one representative sample per class, which reduces the computational burden compared to other machine learning models that need larger training datasets.

\begin{algorithm}[!h]
\caption{Classifying Supraglacial Lakes using RPS-GMM}
\begin{algorithmic}[1]
\label{alg:rps_gmm}
\STATE Set of supraglacial lake classes, $C=\{C_1, C_2, C_3\}=\{refreeze, drain, buried\}$
\STATE \textbf{Input:} Time series data $D$ of backscatter difference $HV_{anom}(t)$ and water percentage $p_{water}(t)$
\STATE \textbf{Output:} Classification of supraglacial lakes into $\{C_1, C_2, C_3\}$

\STATE Define time series of $HV_{anom}(t)$ and $p_{water}(t)$
\STATE Select one representative sample $s_{C_i}$ where $C_i \in C$
\STATE Initialize grid search for time delay $\tau$ and embedding dimension $d$
\FOR {each combination of $\tau$ and $d$}
    \STATE Construct RPS for $s_{C_i}$ where $C_i \in C$ using $\tau$ and $d$
    \STATE Initialize GMM with 10 mixtures
    \STATE Train a set of GMM models $M = \{M_1, M_2, M_3\}$ using the EM algorithm on $s_{C_i}$ where $C_i \in C$
    \FOR {each instance $i \in D$}
        \STATE Construct RPS for $i$ using $\tau$ and $d$
        \STATE Compute likelihood score $S_i$ of $i$ for each $M_j$
        \STATE Assign predicted class based on the maximum likelihood score in $S_i$
    \ENDFOR
    \STATE Calculate and store accuracy for $\tau$ and $d$
\ENDFOR
\STATE Select optimal $\tau^*$ and $d^*$ with the maximum accuracy
\STATE \textbf{Return:} Class labels for $D$ with optimal $\tau^*$ and $d^*$
\end{algorithmic}
\end{algorithm}

During the testing phase, we create the RPS for each instance in the dataset using the current combination of $\tau$ and $d$ in the grid search. Each instance’s RPS is evaluated against the trained GMMs, and the likelihood score is computed for each GMM. Using the maximum likelihood classifier, the class with the highest likelihood score is assigned to the test instance, classifying each supraglacial lake as either \textit{refreeze}, \textit{drain}, or \textit{buried}.

To evaluate the performance of our model, we calculate accuracy for each combination of $\tau$ and $d$ during the grid search. The optimal values $\tau^*$ and $d^*$ are chosen based on the highest accuracy achieved. A step-by-step outline of the proposed methodology is provided in \Cref{alg:rps_gmm}. The dataset and the code are available on GitHub\footnote{\url{https://github.com/ehfahad/TSC-of-Supraglacial-Lakes-Evolution-over-GrIS}}.

\section{Results and Discussion}

This section presents the evaluation results of the RPS-GMM model for classifying supraglacial lakes and compares its performance with several established ML and DL models for time series classification.

\subsection{RPS-GMM Performance Evaluation}

The RPS-GMM model is trained with two sets of features: $HV_{anom}$ alone and a combination of $HV_{anom}$ and $p_{water}$. The reason for having two sets of features is that we aim to evaluate the impact of including the water percentage feature on the model's performance. Our dataset includes time series data from 777 lakes, categorized into 189 refreeze, 392 drained, and 196 buried lakes, across all six subregions of the GrIS for the years 2018 and 2019.

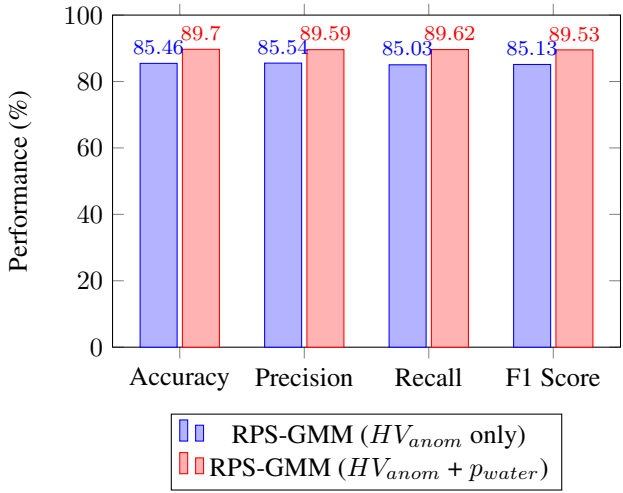
\begin{figure}[!h]
\centering
\begin{tikzpicture}
    \begin{axis}[
        width=\linewidth,  
        height=6cm,  
        ybar,  
        bar width=14pt,  
        symbolic x coords={Accuracy, Precision, Recall, F1 Score},  
        xtick=data,  
        ymin=0, ymax=100,  
        ylabel={Performance (\%)},  
        enlarge x limits=0.18,  
        legend style={at={(0.5,-0.2)},  
        anchor=north,legend columns=1},  
        nodes near coords,  
        nodes near coords align={vertical},  
        every node near coord/.append style={font=\footnotesize, /pgf/number format/fixed},  
        ]
        \addplot coordinates {(Accuracy, 85.46) (Precision, 85.54) (Recall, 85.03) (F1 Score, 85.13)};
        \addplot coordinates {(Accuracy, 89.70) (Precision, 89.59) (Recall, 89.62) (F1 Score, 89.53)};
        \legend{RPS-GMM ($HV_{anom}$ only), RPS-GMM ($HV_{anom}$ + $p_{water}$)}  
    \end{axis}
\end{tikzpicture}
\caption{Performance of the RPS-GMM models}
\label{fig:rps_gmm_results}
\end{figure}

As shown in \Cref{fig:rps_gmm_results}, the model trained with only the backscatter difference achieves an accuracy of 85.46\%. Incorporating the water percentage substantially improves the accuracy to 89.70\%. This improvement underscores the value of adding water percentage with the backscatter signal. Despite an imbalanced dataset, the weighted averages of precision, recall, and F1 score also reflect better performance across all classes when incorporated $p_{water}$ with $HV_{anom}$.

\subsection{Comparison against Existing ML/DL Models}

To further validate the RPS-GMM model, we compare its performance with several established ML and DL models using the sktime time series package \cite{sktime}. Each model is evaluated through $5$-fold cross-validation, with 80\% of the data ($\sim$622 samples) used for training and 20\% for testing in each fold.

\begin{table}[h]
\centering
\caption{Accuracy comparison against ML/DL models}
\label{tab:results_compare}
\begin{tabular}{lcc}
\hline
\textbf{Model} & $\mathbf{HV_{anom}}$ & $\mathbf{HV_{anom}} + \mathbf{p_{water}}$ \\
\hline

LSTMFCNClassifier & 84\% & 87\% \\
FCNClassifier & 51\% & 43\% \\
ResNetClassifier & 57\% & 45\% \\
SimpleRNNClassifier & 50\% & 49\% \\
KNeighborsTimeSeriesClassifier & 80\% & 75\% \\
\textbf{RPS-GMM} & \textbf{85.46\%} & \textbf{89.70\%} \\
\hline
\end{tabular}
\end{table}

The results, detailed in \Cref{tab:results_compare}, show that the RPS-GMM model outperforms all compared ML and DL models in terms of accuracy. For example, the highest accuracy among existing models using only the backscatter difference is 84\%, by the \textit{LSTMFCNClassifier}. When incorporating both features, most models experience decreased performance due to the added complexity and limited training sample size. The \textit{LSTMFCNClassifier} shows a slight improvement, possibly due to its inherent ability to capture long-term temporal dependencies, achieving 87\% accuracy compared to the RPS-GMM model's 89.70\%.

\subsection{Discussion}
The superior performance of the RPS-GMM model, especially when including the water percentage feature, highlights the effectiveness of a comprehensive feature set for accurately classifying supraglacial lakes. Notably, the RPS-GMM model, trained on only one representative sample per class, significantly outperforms models trained on 80\% of the data in $5$-fold cross-validation. This efficiency demonstrates the model's robustness in capturing supraglacial lake dynamics with minimal training data. In contrast, existing ML and DL models showed varying effectiveness, with deep learning models generally underperforming, mainly when both features were included. This underperformance may be due to the complexity of the time series data and the relatively small dataset size, which might not be sufficient for effective training of deep learning models.

\section{Conclusion}

Identifying which supraglacial lakes refreeze, drain, or get buried is crucial for understanding their role in ice sheet dynamics and assessing their impact on global sea level rise. Accurate monitoring of these lakes helps evaluate meltwater dynamics and their implications for the Greenland Ice Sheet (GrIS) mass balance. This study presents a computationally efficient time series classification approach for supraglacial lakes using Reconstructed Phase Space (RPS) and Gaussian Mixture Models (GMM). By integrating time series data of backscatter difference ($HV_{anom}$) from Sentinel-1 and water percentage ($p_{water}$) from Sentinel-2, we demonstrated that incorporating multiple features significantly enhances classification accuracy.

Our results show that the RPS-GMM model, when incorporating both $HV_{anom}$ and $p_{water}$, achieved an accuracy of 89.70\%, outperforming the model trained with only $HV_{anom}$, which attained 85.46\%. This improvement emphasizes the value of combining diverse features to capture the complex dynamics of supraglacial lake evolution. Comparative analysis with established machine learning and deep learning models further underscores the robustness of the RPS-GMM model, which consistently exceeded the performance of these methods. Specifically, while the inclusion of $p_{water}$ improved the RPS-GMM model’s accuracy, the performance of other models declined due to increased complexity and insufficient training data.

The RPS-GMM model demonstrates significant computational efficiency by training on only a single representative sample per class, substantially reducing the computational burden compared to traditional machine learning and deep learning models, which require extensive datasets. This efficiency underscores the model's effectiveness as a tool for monitoring and analyzing supraglacial lakes, which is crucial for understanding meltwater dynamics and their impact on the GrIS's mass balance. To further enhance this methodology, future research could integrate additional features, such as temperature and surface elevation changes, and explore how these features may causally impact the evolution of the supraglacial lakes \cite{hossain2024incorporating}. This could provide deeper insights into polar hydrology and help build more robust and interpretable machine learning models.

\section*{Acknowledgement}

This work is supported by iHARP: NSF HDR Institute for Harnessing Data and Model Revolution in the Polar Regions (Award\# 2118285). The views expressed in this work do not necessarily reflect the policies of the NSF, and endorsement by the Federal Government should not be inferred.

\bibliographystyle{IEEEtran} 


\bibliography{References}

\end{document}